\pdfoutput=1

\documentclass[11pt]{article}

\usepackage[preprint]{coling}

\usepackage{times}
\usepackage{latexsym}

\usepackage[T1]{fontenc}

\usepackage[utf8]{inputenc}

\usepackage{microtype}

\usepackage{inconsolata}

\usepackage{graphicx}
\usepackage{rotating}
\usepackage{adjustbox}
\usepackage{amsmath}

\title{Exploring Large Language Models on Cross-Cultural Values in Connection with Training Methodology}

\author{Minsang Kim \\
  Korea University \\
  Dept. of Computer Science and Engr. \\
  South Korea \\
  \texttt{kmswin1@korea.ac.kr} \\\And
  Seungjun Baek \\
  Korea University \\
  Dept. of Computer Science and Engr. \\
  South Korea \\
  \texttt{sjbaek@korea.ac.kr} \\}

\begin{document}
\maketitle
\begin{abstract}
Large language models (LLMs) closely interact with humans, and thus need an intimate understanding of the cultural values of human society. In this paper, we explore how open-source LLMs make judgments on diverse categories of cultural values across countries, and its relation to training methodology such as model sizes, training corpus, alignment, etc. Our analysis shows that LLMs can judge socio-cultural norms similar to humans but less so on social systems and progress. In addition, LLMs tend to judge cultural values biased toward Western culture, which can be improved with training on the multilingual corpus. We also find that increasing model size helps a better understanding of social values, but smaller models can be enhanced by using synthetic data. Our analysis reveals valuable insights into the design methodology of LLMs in connection with their understanding of cultural values.
\end{abstract}

\section{Introduction \& Related Work}

Large Language Models~(LLMs)  have become a core technology in many real-world applications to closely interact with humans. Thus, it is important that LLMs understand the cultural values and diversities of human societies. 
In particular, since LLMs are widely used as general assistants~\cite{achiam2023gpt, team2023gemini}, they should generate responses which capture the cultural context of target users. For example, people in some countries may think that \texttt{divorce} is largely unacceptable, but those in other countries may consider it as an unpleasant but acceptable practice.

Recently, there have been studies on how language models judge moral norms. \cite{schramowski2022large} investigated whether English language models such as BERT have moral bias like humans in English culture. \cite{ramezani2023knowledge} proposed a methodology for probing moral norms of English language models across diverse cultures. In their experiments, models trained in English predict empirical moral norms across countries worse than the English moral norms. Thus, the authors argue that the moral norms inferred by LLMs have potential risks of bias, and the moral norms suggested by the models should not be regarded as definitive values of morality. 
Recently, \cite{arora2023probing} examined bidirectional language models such as BERT~\cite{devlin2018bert} and RoBERTa~\cite{liu2019roberta} on cross-cultural values in different countries. 
While the authors extensively analyzed the LLMs' predictions, they did not examine the links between training/design choices for LLMs and the human-likeness of LLMs' judgments.

In this paper, we extensively study the progress of LLMs in cultural values across countries using World Value Survey~\cite{Haerpfer2022}. We provide detailed observations on the capability and limitations of LLMs and their relation to the design choices for LLMs. 
We systemically study recent models such as LLaMA~\cite{touvron2023llama, dubey2024llama}, Phi~\cite{abdin2024phi}, and Yi~\cite{young2024yi} on diverse and debatable categories of cultural values. In particular, we discuss how pre-training methodology ~\cite{touvron2023llama, team2024gemma, young2024yi, abdin2024phi} and model attributes such as model sizes, training corpus, alignment, etc., affect the LLMs' understanding of human cultures. We expect that our findings provide key insights on design methodology to improve the quality of cultural judgments by LLMs across countries.

Our findings are summarized as follows: 
i) LLMs similarly judge Socio-Cultural Norms with humans, but are less similar in Social Systems and Progress. ii) LLMs' assessment of cultural values tends to be biased toward Western cultures. iii) However, cultural diversity can be improved by training on the multilingual corpus. iv) Larger models have a stronger cultural-awareness. v) Synthetic data can distill cultural knowledge of larger language models to smaller models. vi) Alignment causes LLMs to resemble humans.

\section{Methodology of probing LLMs}

\subsection{Dataset - World Value Survey} \label{sec:wvs}

We use the World Value Survey (WVS)~\cite{Haerpfer2022} dataset to probe LLMs's knowledge of cultural values across countries. WVS contains questions on diverse topics of social values. There are 209 questions from 12 categories in 55 countries. To probe LLM, we convert all questions and answers to multiple-choice tasks using the template in  Fig.~\ref{fig:prompt_wvs} in Appendix~\ref{appendix:prompt}. Each question has $K$ candidate answers $a_k$. A sample question: \texttt{``Is the death penalty justifiable?''} has $K=3$ candidate answers: \texttt{\{A) Never Justifiable, B) Neutral C) Justifiable\}}. 
In our experiment, $K$ takes values from $\{3, 5, 7, 9\}$. Table~\ref{tab:dataset_wvs} in Appendix~\ref{appendix:dataset} shows dataset statistics of WVS.

\subsection{Method - Scoring Cultural Values}

Below we describe a method to measure the correlation between the answers provided by LLM and humans. We individually assign scores to candidate answers which enables quantifying correlations in the choice of answers. 
Let $s_k$ denote the score assigned to $k$-th candidate answer for $k\in[K]$. We let $s_k$ take values between -1 and 1. Since there are $K$ answers, we regularly distribute $s_1,\dots,s_K$ over interval $[-1,1]$. For example, the question \texttt{``Is the death penalty justifiable?''} has candidate answers \texttt{never justifiable}, \texttt{neutral}, or \texttt{justifiable}.
We assign the scores -1, 0, and 1 to the candidate answers. For each question, We will compute the \emph{mean score} which is the expectation of score over the distribution of answers generated by LLMs and humans, and compare the correlations of their average scores.

Firstly, we define the distribution of answer candidates chosen by LLMs. Let $p_k$ denote the probability of generating $k$-th answer candidate for $k\in[K]$ conditional on the question. 
 $p_k$ is defined to be the likelihood (softmax output) of answer $k$ of LLM model $P$: 
\begin{equation}
    p_{k} = \prod_{i=c+1}^{c+l_k} P({w}_i | {w}_{i-1},...,{w}_{1} )
\end{equation}
where $w_i$ denotes the $i$-th token, $l_k$ denotes token length of $k$-th answer candidate, and $c$ denotes the length of question contexts. We normalize $p_k$'s to yield a probability distribution \(
    \hat{p}_{k} = p_k/(\sum_{i=1}^{K} p_i)
\)
Thus, the mean score of LLMs is given by $  \sum_{i=1}^K \hat p_k s_k$.

Secondly, we compute the distribution of answer choices by humans. The distribution denoted by $r_k$, $k\in [K]$, is simply set to the empirical distribution of choosing answer candidate $k$ for the given question. The statistics of answer distribution by the survey participants are available in the WVS dataset. Thus, the mean score of humans is given by $  \sum_{i=1}^K r_k s_k$.

Finally, we compute the Pearson's correlation~\cite{freedman2007statistics} between mean scores over questions to measure the degree of agreement between  LLM and humans.



\begin{figure*}[t!]
    \centering
    \includegraphics[width=1.\textwidth]{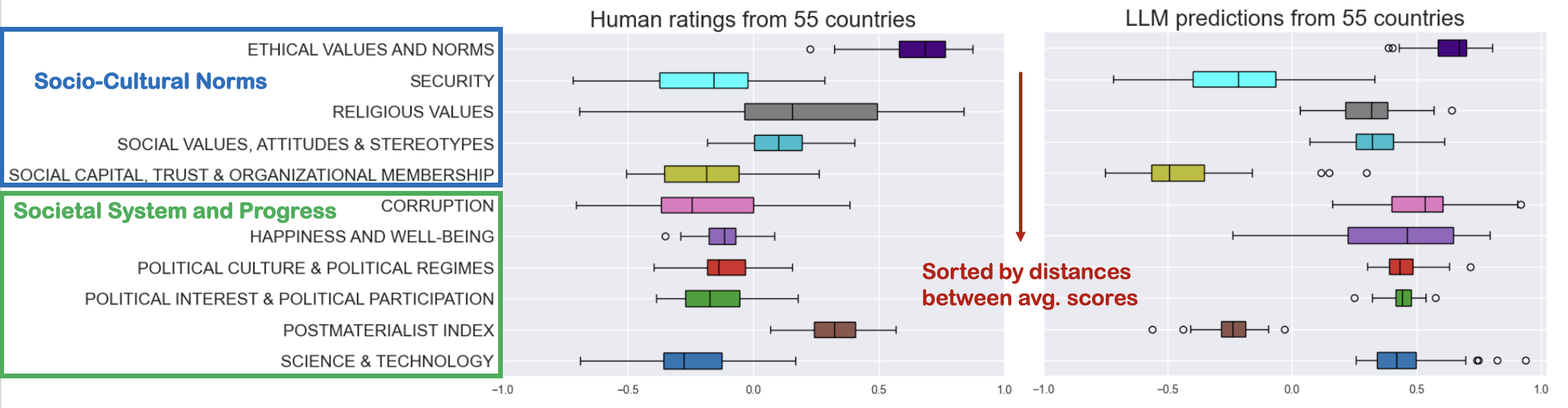}
    \caption{Comparison of human-ratings and LLM predictions across 55 countries in WVS~\cite{Haerpfer2022}. Left: Boxplots of human ratings per category across countries. Right: Corresponding cultural judgment scores estimated by Llama-2-70b Chat~\cite{touvron2023llama}.}
    \label{fig:corr-boxplot}
\end{figure*}

\begin{figure*}[t!]
    \centering
    \includegraphics[width=1.\textwidth]{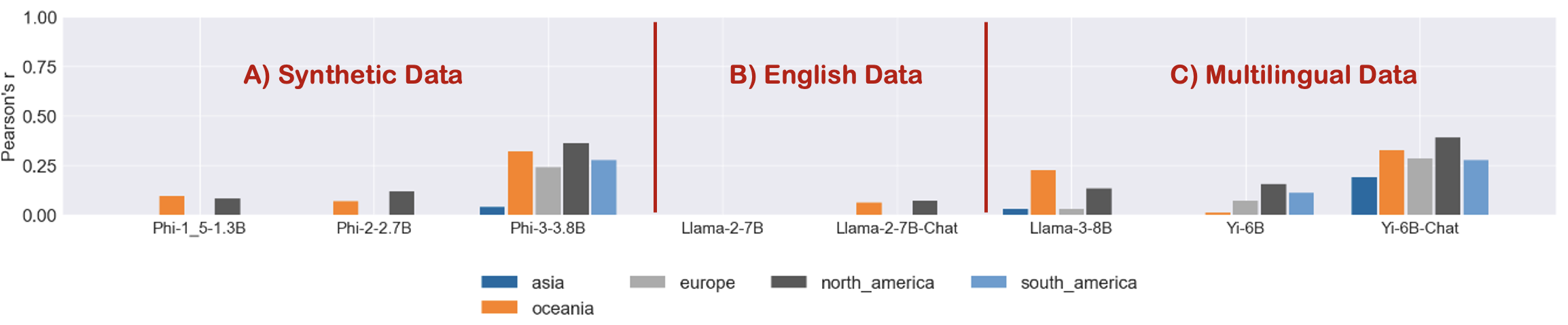}
    \caption{Average Pearson's correlation of small sizes model up to 8B across all countries grouped by continents.}
    \label{fig:corr-continent}
\end{figure*}

\begin{figure*}[t!]
    \centering
    \includegraphics[width=1.\textwidth]{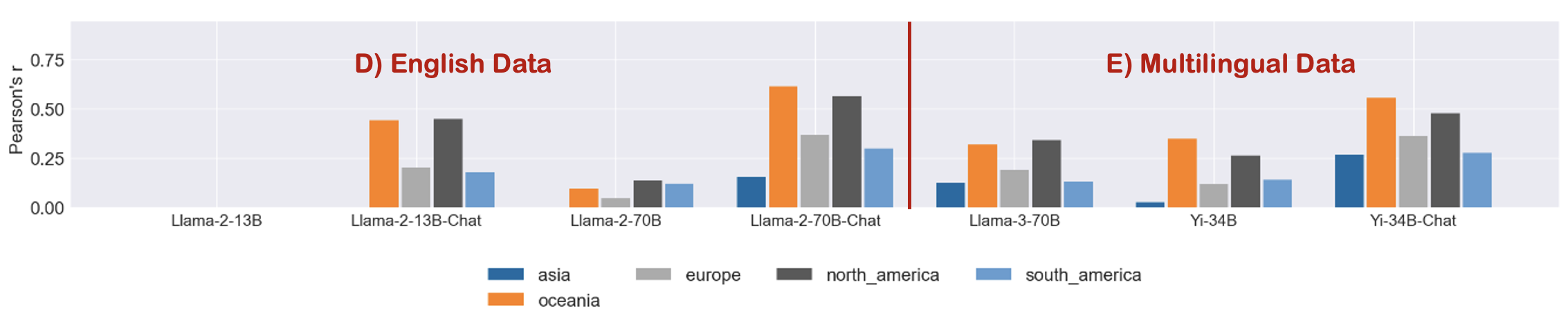}
    \caption{Average Pearson's correlation of larger sizes model including 13B, 34B, and 70B across all countries grouped by continents.}
    \label{fig:corr-continent-large}
\end{figure*}

\section{Experiment \& Analysis}

\subsection{Experimental settings}
\textbf{Baselines.} We evaluate various open-source LLMs such as phi-series~\cite{li2023textbooks, gunasekar2023textbooks,  abdin2024phi}, Yi~\cite{young2024yi}, Llama-2~\cite{touvron2023llama}, and Llama-3~\cite{dubey2024llama}. 
We chose the open models because the model information such as sizes and training tokens are publicly available, which enables detailed analysis.
The models are categorized into small or large models. The numbers of parameters for small models are between 1B and 8B parameters; for large models, the sizes are 13B, 34B, and 70B. If a chat model associated with the base model is available, we include it in the analysis to examine the effects of alignment. Table~\ref{tab:training_tokens} in appendix~\ref{appendix:probing} summarizes the model and pre-training dataset sizes for analysis. 



\subsection{Exploring LLMs' Judgments on WVS and connections to Training Methodology}\label{sec:corr}



We examine how similar the LLMs' inference to that of humans in various categories of cultural values, and its relation to training methods. We aggregate the mean scores and compute their correlations over WVS dataset to measure the similarity.

\noindent \textbf{Obs.\ 1. LLMs and Humans judge similarly on Socio-Cultural Norms, but less so on Social Systems and Progress.} Fig.~\ref{fig:corr-boxplot} show the ratings by LLMs and humans across various topics in descending order of similarity between LLMs and humans. We roughly divide topics into two topic groups based on similarity: we observe that LLMs tend to agree with humans on \emph{social-cultural norms} group, but less so on \emph{social systems and progress} group. Since LLMs are trained from massive web data, they can understand socio-cultural norms such as ethical values, security, social attitudes, etc.
For example, a question in \textbf{Ethical values} is \texttt{Is cheating on taxes justifiable?}, which seems commonsense and easy to agree on across cultures. 
By contrast, LLMs diverge from humans in \emph{social systems and progress}. A question in \textbf{Science \& Technology,}  topic is  \texttt{Does technology make our lives better?}. The issue is controversial and subtle across culture, e.g., some may believe the flourishing technology in AI may be eventually harmful to mankind. Also, the topics on social systems include questions such as \texttt{Is democratic political system desirable?}, and \texttt{What is more important for better country: economic growth or strong military force?} It appears to be still challenging for LLMs to comprehend the variations in cultural contexts and nuances across countries in its entirety. 
This shows the importance of recognizing the limitations and potential risks of using LLMs, and of managing them accordingly.

\begin{figure*}[t!]
    \centering
    \includegraphics[width=1.\textwidth]{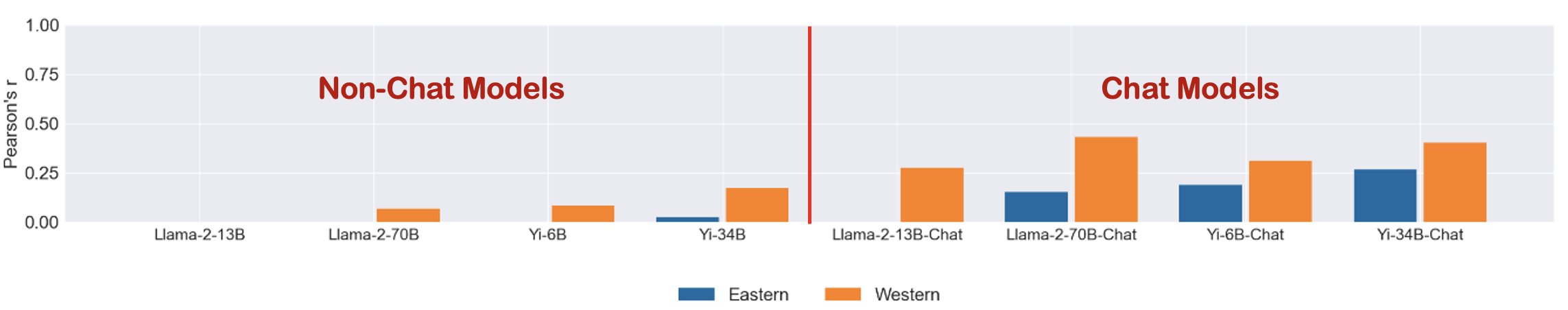}
    \caption{Comparison average Pearson's correlation between Chat models Vs. Non-Chat models.}
    \label{fig:chat-comparison}
\end{figure*}

\noindent \textbf{Obs.\ 2. Cultural Views of LLMs match better with the Western World.} As shown in  Fig.~\ref{fig:corr-continent} and Fig.~\ref{fig:corr-continent-large}, the LLM's view on cultural values is more similar to Western than non-Western countries.
Examples of Western countries are the USA, Canada, Australia, and Germany. LLMs exhibit a low correlation with cultural values of the non-Western world, particularly  Asia. These findings indicate that most LLMs remain biased towards Western cultural values due to their training on extensive English-language datasets which are likely to contain a greater representation of knowledge in Western cultures.

\noindent \textbf{Obs.\ 3. But, Cultural Diversity can be better captured by training on Multilingual Corpus.}
While most open-source LLMs exhibit a bias toward Western cultures,  Yi models \cite{young2024yi} show significantly higher correlations with non-Western cultures. Yi has been trained on a substantial corpus of both English and Chinese data. This suggests that incorporating a large Chinese corpus can markedly enhance performance in non-Western cultural contexts. Furthermore, Llama-3-7B and Llama-3-70B, which are trained on over 5\% high-quality non-English data encompassing more than 30 languages, show notably improved correlations with Asian (Eastern) cultures compared to  Llama-2 (non-chat models). Interestingly, our results suggest that LLMs may acquire cross-cultural knowledge through training on multilingual data, and this knowledge may be transferred across different languages, where a similar observation was made in~\cite{zhao2024large}.
 
\noindent \textbf{Obs. 4. Larger Models have Better Insights on Cultural Values as well.} Even though reasoning and knowledge capabilities are typically known to improve with model size, the effects of model size on cultural judgments have been less examined. By comparing Fig.~\ref{fig:corr-continent} and Fig.~\ref{fig:corr-continent-large}, we observe that larger models exhibit a higher correlation with human ratings.
For example, while Llama-2-7B-Chat struggles in capturing cultural values like humans, Llama-2-13B-Chat and Llama-2-70B-Chat achieve increasingly higher correlation with model size. Similarly, Yi-34B-Chat achieves a significantly higher correlation than Yi-6B-Chat. Our findings underscore that model size is crucial for learning human cultural values as well.

\noindent \textbf{Obs. 5. But, Models may overcome Size Limit with Synthetic Data.}
Although the model size is a large factor, training with \emph{synthetic data} may boost the performance of smaller models. In Fig.~\ref{fig:corr-continent}, phi-3-\textbf{3.8B} trained on \texttt{A) Synthetic data} achieves higher correlation than Llama-3-\textbf{8B} trained on \texttt{C) Multilingual data}, or chat-model Llama-2-\textbf{7B-chat} trained on \texttt{B) English data} (chat models are considered as advantageous to base models).
Although details of synthetic data for phi ~\cite{abdin2024phi} are not disclosed, it is our guess that synthetic data is generated by larger models~\cite{benallal2024cosmopedia}. Our results show that such synthetic data may significantly differ from web data from the viewpoint of social science. Our interpretation is that, each piece of data created by large models may have a culturally unified view because of its massive cross-cultural knowledge, whereas individual web data is invariably limited by the subjectivity and cultural bias of the human author.

\noindent \textbf{Obs. 6. Alignment Enhances Model's Human-Likeness in Cultural Views.} Fig.~\ref{fig:chat-comparison} shows the comparison between non-chat models and chat models from the same base model. We observe that current alignment techniques successfully adjust LLMs to make social judgments similar to humans. For instance, while Llama-2-13B and Llama-2-70B without alignment show a low correlation with human judgments, both models show much higher correlations when fine-tuned as chat models. This trend is observed in other models. Yi-6B and Yi-34B achieve significantly higher correlations after alignment. Thus, we find that alignment, by which LLMs learn human preferences, also enables LLMs to make judgments on cultural values like humans.

\section{Conclusion}
In this paper, we extensively examine how LLMs make judgments of cultural values. In our experiments, current LLMs have limitations of cross-cultural judgments specific to questions categorized in societal systems and progress and tend to be biased toward Western culture. However, we also find that training multilingual data can transfer cross-cultural knowledge to English. In addition, since larger models have a better perception of cultural values, training synthetic data created by larger models can be helpful. Finally, alignment successfully trains the model of human likeness.



\section{Limitations}
Despite our extensive study of various language models (LLMs) across diverse categories of cultural values, our research remains confined to the World Value Survey dataset. Consequently, we plan to further investigate the properties of LLMs' cultural judgment using other datasets in the future. Furthermore, due to the lack of technical details~\cite{li2023textbooks, gunasekar2023textbooks, abdin2024phi}, we conduct a preliminary analysis of the properties of synthetic data. Although we indirectly present our rationale for this analysis as derived from other well-known synthetic data~\cite{benallal2024cosmopedia}, we aim to provide a comprehensive overview.

\bibliography{custom}

\begin{thebibliography}{18}
\providecommand{\natexlab}[1]{#1}

\bibitem[{Abdin et~al.(2024)Abdin, Jacobs, Awan, Aneja, Awadallah, Awadalla, Bach, Bahree, Bakhtiari, Behl et~al.}]{abdin2024phi}
Marah Abdin, Sam~Ade Jacobs, Ammar~Ahmad Awan, Jyoti Aneja, Ahmed Awadallah, Hany Awadalla, Nguyen Bach, Amit Bahree, Arash Bakhtiari, Harkirat Behl, et~al. 2024.
\newblock Phi-3 technical report: A highly capable language model locally on your phone.
\newblock \emph{arXiv preprint arXiv:2404.14219}.

\bibitem[{Achiam et~al.(2023)Achiam, Adler, Agarwal, Ahmad, Akkaya, Aleman, Almeida, Altenschmidt, Altman, Anadkat et~al.}]{achiam2023gpt}
Josh Achiam, Steven Adler, Sandhini Agarwal, Lama Ahmad, Ilge Akkaya, Florencia~Leoni Aleman, Diogo Almeida, Janko Altenschmidt, Sam Altman, Shyamal Anadkat, et~al. 2023.
\newblock Gpt-4 technical report.
\newblock \emph{arXiv preprint arXiv:2303.08774}.

\bibitem[{Arora et~al.(2023)Arora, Kaffee, and Augenstein}]{arora2023probing}
Arnav Arora, Lucie-Aim{\'e}e Kaffee, and Isabelle Augenstein. 2023.
\newblock Probing pre-trained language models for cross-cultural differences in values.
\newblock In \emph{Proceedings of the First Workshop on Cross-Cultural Considerations in NLP (C3NLP)}, pages 114--130.

\bibitem[{Ben~Allal et~al.(2024)Ben~Allal, Lozhkov, Penedo, Wolf, and von Werra}]{benallal2024cosmopedia}
Loubna Ben~Allal, Anton Lozhkov, Guilherme Penedo, Thomas Wolf, and Leandro von Werra. 2024.
\newblock \href {https://huggingface.co/datasets/HuggingFaceTB/cosmopedia} {Cosmopedia}.

\bibitem[{Devlin(2018)}]{devlin2018bert}
Jacob Devlin. 2018.
\newblock Bert: Pre-training of deep bidirectional transformers for language understanding.
\newblock \emph{arXiv preprint arXiv:1810.04805}.

\bibitem[{Dubey et~al.(2024)Dubey, Jauhri, Pandey, Kadian, Al-Dahle, Letman, Mathur, Schelten, Yang, Fan et~al.}]{dubey2024llama}
Abhimanyu Dubey, Abhinav Jauhri, Abhinav Pandey, Abhishek Kadian, Ahmad Al-Dahle, Aiesha Letman, Akhil Mathur, Alan Schelten, Amy Yang, Angela Fan, et~al. 2024.
\newblock The llama 3 herd of models.
\newblock \emph{arXiv preprint arXiv:2407.21783}.

\bibitem[{Freedman et~al.(2007)Freedman, Pisani, and Purves}]{freedman2007statistics}
David Freedman, Robert Pisani, and Roger Purves. 2007.
\newblock Statistics (international student edition).
\newblock \emph{Pisani, R. Purves, 4th edn. WW Norton \& Company, New York}.

\bibitem[{Gunasekar et~al.(2023)Gunasekar, Zhang, Aneja, Mendes, Del~Giorno, Gopi, Javaheripi, Kauffmann, de~Rosa, Saarikivi et~al.}]{gunasekar2023textbooks}
Suriya Gunasekar, Yi~Zhang, Jyoti Aneja, Caio C{\'e}sar~Teodoro Mendes, Allie Del~Giorno, Sivakanth Gopi, Mojan Javaheripi, Piero Kauffmann, Gustavo de~Rosa, Olli Saarikivi, et~al. 2023.
\newblock Textbooks are all you need.
\newblock \emph{arXiv preprint arXiv:2306.11644}.

\bibitem[{Haerpfer et~al.(2022)Haerpfer, Inglehart, Moreno, Welzel, Kizilova, Diez-Medrano, Lagos, Norris, and Puranen}]{Haerpfer2022}
Christian Haerpfer, Ronald Inglehart, Alejandro Moreno, Christian Welzel, Kseniya Kizilova, Jaime Diez-Medrano, Marta Lagos, Pippa Norris, and E~Ponarin \&~B Puranen. 2022.
\newblock \href {https://doi.org/doi:10.14281/18241.20} {World values survey: Round seven – country-pooled datafile version 5.0.0.}
\newblock \emph{Madrid, Spain \& Vienna, Austria: JD Systems Institute \& WVSA Secretariat.}

\bibitem[{Li et~al.(2023)Li, Bubeck, Eldan, Del~Giorno, Gunasekar, and Lee}]{li2023textbooks}
Yuanzhi Li, S{\'e}bastien Bubeck, Ronen Eldan, Allie Del~Giorno, Suriya Gunasekar, and Yin~Tat Lee. 2023.
\newblock Textbooks are all you need ii: phi-1.5 technical report.
\newblock \emph{arXiv preprint arXiv:2309.05463}.

\bibitem[{Liu(2019)}]{liu2019roberta}
Y~Liu. 2019.
\newblock Roberta: A robustly optimized bert pretraining approach.
\newblock \emph{arXiv preprint arXiv:1907.11692}.

\bibitem[{Ramezani and Xu(2023)}]{ramezani2023knowledge}
Aida Ramezani and Yang Xu. 2023.
\newblock Knowledge of cultural moral norms in large language models.
\newblock In \emph{The 61st Annual Meeting Of The Association For Computational Linguistics}.

\bibitem[{Schramowski et~al.(2022)Schramowski, Turan, Andersen, Rothkopf, and Kersting}]{schramowski2022large}
Patrick Schramowski, Cigdem Turan, Nico Andersen, Constantin~A Rothkopf, and Kristian Kersting. 2022.
\newblock Large pre-trained language models contain human-like biases of what is right and wrong to do.
\newblock \emph{Nature Machine Intelligence}, 4(3):258--268.

\bibitem[{Team et~al.(2023)Team, Anil, Borgeaud, Wu, Alayrac, Yu, Soricut, Schalkwyk, Dai, Hauth et~al.}]{team2023gemini}
Gemini Team, Rohan Anil, Sebastian Borgeaud, Yonghui Wu, Jean-Baptiste Alayrac, Jiahui Yu, Radu Soricut, Johan Schalkwyk, Andrew~M Dai, Anja Hauth, et~al. 2023.
\newblock Gemini: a family of highly capable multimodal models.
\newblock \emph{arXiv preprint arXiv:2312.11805}.

\bibitem[{Team et~al.(2024)Team, Mesnard, Hardin, Dadashi, Bhupatiraju, Pathak, Sifre, Rivi{\`e}re, Kale, Love et~al.}]{team2024gemma}
Gemma Team, Thomas Mesnard, Cassidy Hardin, Robert Dadashi, Surya Bhupatiraju, Shreya Pathak, Laurent Sifre, Morgane Rivi{\`e}re, Mihir~Sanjay Kale, Juliette Love, et~al. 2024.
\newblock Gemma: Open models based on gemini research and technology.
\newblock \emph{arXiv preprint arXiv:2403.08295}.

\bibitem[{Touvron et~al.(2023)Touvron, Martin, Stone, Albert, Almahairi, Babaei, Bashlykov, Batra, Bhargava, Bhosale et~al.}]{touvron2023llama}
Hugo Touvron, Louis Martin, Kevin Stone, Peter Albert, Amjad Almahairi, Yasmine Babaei, Nikolay Bashlykov, Soumya Batra, Prajjwal Bhargava, Shruti Bhosale, et~al. 2023.
\newblock Llama 2: Open foundation and fine-tuned chat models.
\newblock \emph{arXiv preprint arXiv:2307.09288}.

\bibitem[{Young et~al.(2024)Young, Chen, Li, Huang, Zhang, Zhang, Li, Zhu, Chen, Chang et~al.}]{young2024yi}
Alex Young, Bei Chen, Chao Li, Chengen Huang, Ge~Zhang, Guanwei Zhang, Heng Li, Jiangcheng Zhu, Jianqun Chen, Jing Chang, et~al. 2024.
\newblock Yi: Open foundation models by 01. ai.
\newblock \emph{arXiv preprint arXiv:2403.04652}.

\bibitem[{Zhao et~al.(2024)Zhao, Zhang, Chen, Kawaguchi, and Bing}]{zhao2024large}
Yiran Zhao, Wenxuan Zhang, Guizhen Chen, Kenji Kawaguchi, and Lidong Bing. 2024.
\newblock How do large language models handle multilingualism?
\newblock \emph{arXiv preprint arXiv:2402.18815}.

\end{thebibliography}

\appendix

\clearpage
\onecolumn

\section{Appendix}
\subsection{Prompt Templates}\label{appendix:prompt}
\begin{figure}[h!]
    \centering
    \fbox{
    \begin{minipage}{17em}
    Question: \{Question\} \\
    Choices: \{A) $a_1$, B) $a_2$ ...\} \\
    In \{Country\}, Answer:
    \end{minipage}
    }
    \caption{Prompt template for WVS questions. For chat models, we apply chat templates in each model.}
    \label{fig:prompt_wvs}
\end{figure}

\subsection{Probing LLMs}\label{appendix:probing}
\begin{table}[h!]
\centering 
\begin{adjustbox}{width=.5\textwidth, center}
\begin{tabular}{|l|l|l|}
\hline
Model   & \# Training tokens & Size         \\ \hline
phi-1.5  & 150B               & 1.3B         \\ 
phi-2   & 1.4T               & 2.7B         \\
phi-3 &   3.3T          &  3.8B      \\
        \\
Llama-2 & 2T                 & 7B, \underline{\textbf{13B, 70B}} \\
Llama-3 & 15T                 & 8B, \underline{\textbf{70B}} \\
Yi      & 3T                 & 6B, \underline{\textbf{34B}}      \\ \hline
\end{tabular}
\end{adjustbox}
\caption{Training tokens and model sizes of models. The \textbf{\underline{Bold with underline}} denotes the group of large models.}
\label{tab:training_tokens}
\end{table}

\subsection{Dataset statistics} \label{appendix:dataset}
\begin{table}[h!]
\begin{adjustbox}{width=.7\textwidth, center}

\begin{tabular}{|l|l|}
\hline
\textbf{Category}                                           & \textbf{\# Questions} \\ \hline
SOCIAL VALUES, ATTITUDES \& STEREOTYPES            & 44           \\ 
HAPPINESS AND WELL-BEING                           & 5            \\ 
SOCIAL CAPITAL, TRUST \& ORGANIZATIONAL MEMBERSHIP & 47           \\ 
CORRUPTION                                         & 5            \\ 
MIGRATION                                          & 8            \\ 
SECURITY                                           & 12           \\ 
POSTMATERIALIST INDEX                              & 3            \\ 
SCIENCE \& TECHNOLOGY                              & 5            \\ 
RELIGIOUS VALUES                                   & 11           \\ 
ETHICAL VALUES AND NORMS                           & 22           \\ 
POLITICAL INTEREST \& POLITICAL PARTICIPATION      & 31           \\ 
POLITICAL CULTURE \& POLITICAL REGIMES             & 16           \\ \hline
TOTAL & 209 \\ \hline
\end{tabular}

\end{adjustbox}
    \caption{\# Dataset per category by World Value Survey.}
    \label{tab:dataset_wvs}
\end{table}

\section{Case Study}

\begin{table}[h!]
\begin{adjustbox}{width=1.1\textwidth, center}
\begin{tabular}{|l|l|l|l|l|}
\hline
\textbf{Category} & \textbf{Question} & \textbf{LLaMA-70B-Chat} & \textbf{ 
Humans} & \textbf{Country}\\\hline
Ethical values \& norms & Can cheating on taxes be justifiable? & Never justifiable (97\%) & Never justifiable (82\%) & USA \\
Security & Are you worried about terrorist attacks? & Not at all (73\%) & Not at all (81\%) & Canada \\
Religious values & Do you believe in the existence of god? & Yes (79\%) & Yes (76\%) & USA \\
Social values \& stereotypes & Is politics important in your life? & Yes (70\%) & Yes (73\%) & France \\
Social capital, trust & Do you believe people of another religion? & Yes (63\%) & Yes (60\%) & Germany \\\hline\hline
Migration & Is immigration Increase the risks of terrorism? & Agree (61\%) & Disagree (69\%) & Australia \\
Postmaterialist index & Which is the most important? & A stable economy (59\%) & Fight against crime (54\%) & Australia \\
Science \& Technology & We depend too much on science and not enough on faith.
 & Agree (65\%) & Disagree (67\%) &  Greece \\
Political culture & a strong leader who does not bother with parliament and elections good or bad? & Good (67\%) & Good (88\%) & Russia \\\hline

\end{tabular}
\end{adjustbox}
    \caption{Predictions of LLaMA2-70B-Chat and Humans on questions of various categories in Western Countries.}
    \label{tab:case_western}
\end{table}

\subsection{Answer examples in Western Cultures}
Table~\ref{tab:case_western} shows the questions and answers from LLM and humans in Western cultures. Results show that LLaMA2-70B-Chat predicts cultural values similar to Western people in socio-cultural categories (top partition in the table) showing a high correlation. By contrast, they can judge cultural values differently from humans in societal systems and progress (bottom partition in the table), which are more complex and debatable depending on cultures.

\begin{table}[h!]
\begin{adjustbox}{width=1.1\textwidth, center}

\begin{tabular}{|l|l|l|l|l|}
\hline
\textbf{Category} & \textbf{Question} & \textbf{Humans} & \textbf{LLaMA2-70B-Chat} & \textbf{Country} \\\hline
Ethical values \& norms & Can sex before marriage be justifiable? & Never justifiable (85\%) & Justifiable (65\%) & South Korea \\
Security & Are you worried about a war involving your country? & Very Much (61\%) & Not at all (70\%) & South Korea \\
Religious values & Do you believe in heaven? & No (68\%) & Yes (84.8\%) & Japan \\
Social values \& stereotypes & Is religion important in your life? & Not important (64\%) & Important (62\%) & Japan \\
Social capital, trust &  Do you believe people of another religion? & Do not trust (80\%) & Trust (64\%) & China \\\hline\hline
Migration & Is immigration Increase the risks of terrorism? & Agree (88\%) & Disagree (62\%) & China \\
Postmaterialist indes & Which is the most important? & A stable economy (63\%) & Fight against crime (86\%) & Singapore \\
Science \& technology & We depend too much on science and not enough on faith. & Agree (61\%) & Disagree (58\%) & Kenya \\
Political culture & a strong leader who does not bother with parliament and elections good or bad? & Bad (60\%) & Good (86\%) & China  \\\hline

\end{tabular}

\end{adjustbox}
    \caption{Predictions of LLaMA-70B-Chat and Humans on questions of various categories in Non-western Countries.}
    \label{tab:case_not_western}
\end{table}

\subsection{Answer examples in Non-western Cultures}
By contrast, Table~\ref{tab:case_not_western} illustrates the questions and answers from LLM and humans in non-Western cultures. Results show that cultural values predicted by LLaMA2-70B-Chat tend to show a low correlation with non-Western people.

\end{document}